\documentclass[nonacm, sigconf]{acmart}
\usepackage[flushleft]{threeparttable}
\usepackage{hyperref}
\usepackage{tabularx}

\AtBeginDocument{%
  }

\begin{document}

\title{NEAT Algorithm-based Stock Trading Strategy with Multiple Technical Indicators Resonance}

\author{LICHUN HUANG}
\email{leo900527@gmail.com}
\affiliation{%
  \institution{Department of Computer Science, National Yang Ming Chiao Tung University }
  \streetaddress{No. 1001, Daxue Rd., East Dist.}
  \city{Hsinchu City}
  \state{Taiwan}
  \country{Republic of China}
  \postcode{300093}
}


\begin{abstract}
In this study, we applied the NEAT (NeuroEvolution of Augmenting Topologies) algorithm to stock trading using multiple technical indicators. Our approach focused on maximizing earning, avoiding risk, and outperforming the Buy \& Hold strategy. We used progressive training data and a multi-objective fitness function to guide the evolution of the population towards these objectives. The results of our study showed that the NEAT model achieved similar returns to the Buy \& Hold strategy, but with lower risk exposure and greater stability. We also identified some challenges in the training process, including the presence of a large number of unused nodes and connections in the model architecture. In future work, it may be worthwhile to explore ways to improve the NEAT algorithm and apply it to shorter interval data in order to assess the potential impact on performance.

\end{abstract}




\maketitle

\section{Introduction}
Predicting stock market trends is a key goal for investors seeking to maximize returns and minimize risk. While there are two main approaches to analyzing stocks and companies (fundamental analysis and technical analysis), technical analysis is particularly popular due to its focus on using various technical indicators to interpret and simplify market data. These indicators are designed to help investors make informed trading decisions. However, accurately predicting stock trends remains challenging due to the complexity and uncertainty of financial markets.

In this project, we propose to use evolutionary algorithms and multi-indicator resonance\cite{resonance} to develop a trading system that is able to navigate these uncertainties and identify profitable opportunities. Specifically, we apply the neuroevolution of augmenting topologies (NEAT) algorithm\cite{1004508}, which is a form of evolutionary computation that allows for the evolution of complex neural network structures through the generation and evaluation of genetically diverse populations of networks. NEAT has been applied successfully to a variety of tasks, including image classification\cite{neat-image} and game playing\cite{neat-game}\cite{neat-game2}. In this paper, we propose a NEAT-based approach for stock trading that aims to optimize multiple objectives, including earning stabilization, capacity utilization minimization, risk avoidance, and benefit maximization.

To evaluate the performance of our NEAT-based approach, we use 22 years of historical stock trading data (2000-2022) for the 503 constituents of the S\&P 500 index. The features for model input are derived from seven technical indicators, including the Simple Moving Average (SMA), Stochastic Oscillator (KD), Moving Average Convergence \& Divergence (MACD), Commodity Channel Index (CCI), Williams \%R, Relative Strength Index (RSI), and Chaikin A/D Oscillator (ADOSC), extracted from this data. We compare the performance of our NEAT-based approach to a buy and hold strategy, which serves as a baseline for comparison. Our goal is to demonstrate that our model can outperform the buy and hold strategy and has the ability to manage capacity, risk, and earn in any transaction targets.

\section{Methodology}
\subsection{Research data}
The dataset used in this project sources from yahoo finance API, which consists of $22$ years of historical stock trading data for the 503 constituents of the S\&P 500 index, covering the period from 1999/12/31 to 2022/11/27. The data was obtained from yahoo finance API sources and has a 1-day interval for each row, resulting in a dataset with 2580890 rows. The dataset includes eight features: Ticker, Datetime, Open, High, Low, Close, Adjusted Close Price, and Volume.\ref{tab:column} This results in a total complexity of $8$ columns x $2580890$ rows  $=20,647,120$.

\begin{table*}
  \caption{Columns of dataset}
  \label{tab:column}
  \begin{tabular}{ccl}
    \toprule
    Column Name & Type     &Meaning\\
    \midrule
     Ticker      & String   & The ticker of the company.              \\
     Datetime    & Datetime & The time of the data.                   \\
     Open        & Float    & The price of open.                      \\
     High        & Float    & The highest price in the time interval. \\
     Low         & Float    & The lowest price in the time interval.  \\
     Close       & Float    & The price of close.                     \\
     Adj\_Close   &Float     &The price of close after the adjustment of dividend and interest. \\
  \bottomrule
\end{tabular}
\end{table*}

\subsection{Model Design}
The input and output of the model are important factors that influence its behavior and performance. In this project, we aim to build a trading bot based on multiple indicators resonance, which requires a large number of technical indicators as input to the model. These indicators are used to capture the complex dynamics of the financial market and provide insight into potential trading opportunities. In addition to these indicators, we also include information about the position the model is holding to enhance its capacity management capabilities.

\begin{figure}
    \centering
    \includegraphics[height=!, width=\linewidth, keepaspectratio]{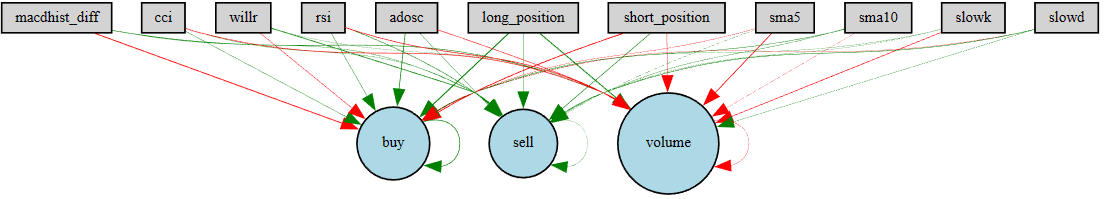}
    \caption{The sample initialized model in the population.}
    \label{fig: initial_structure}
\end{figure}

In NEAT algorithm, it is a common practice to start with a relatively simple model and allow it to evolve and become more complex overtime. This can help to reduce the risk of overfitting and improve the overall performance of the model. In this project, we initialize the population with fully connected recurrent neural networks (RNNs) like the structure shown in \ref{fig: initial_structure}. RNNs are a type of neural network that are particularly well-suited for processing sequential data, such as time series data. They are able to capture dependencies between elements in the sequence, which can be useful in tasks such as stock prediction. By starting with a fully connected RNN, you are giving the model the ability to learn and adapt to the data as it evolves over time. As the model evolves, it may choose to add or remove connections between neurons in order to better fit the data and improve its performance.
\subsubsection{Input }
The input of the model is shown in \ref{tab: input}. The \emph{long\_position} and \emph{short\_position} represent the proportions of the model's total assets that are currently held in long and short positions, respectively. The \emph{sma5} and \emph{sma10} are simple moving averages calculated over the past 5 and 10 days, respectively, and are divided by the most recent closing price for normalization. These indicators are used to smooth out price data and identify trends. The \emph{slow\_k} and \emph{slow\_d} are Stochastic Oscillators, which are technical indicators used to measure the strength of a trend and identify overbought and oversold conditions. The \emph{willr} (Larry Williams' R\%) is a technical indicator that is used to measure the strength of a trend. The \emph{macd\_diff} is the difference between the current and previous values of the Moving Average Convergence Divergence (MACD) indicator, normalized by the current closing price. The MACD is used to identify the direction and strength of a trend. The \emph{cci} (Commodity Channel Index) and \emph{rsi} (Relative Strength Index) are technical indicators used to identify overbought and oversold conditions. The \emph{adosc} (Accumulation/Distribution Oscillator) is a technical indicator that measures the trend of price and volume. These indicators can be useful in analyzing the performance of a stock and making informed trading decisions.

\begin{table*}
  \caption{Selected inputs.}
  \label{tab: input}
  \begin{center}

  \begin{tabular}{ccl}
    \toprule
    Symbol         & Indicators     & Formula\\
    \midrule
    \emph{long\_position}  & Long Position  & $P_{long}/A$\\
    \emph{short\_position}  & Short Position & $P_{short}/A$\\
    \emph{sma5}            & Simple $n$(5 \emph{here})-days Moving Average    & $C_t/\frac{C_t+C_{t-1}+...+C_{t-n}}{n}, n=5$\\
    \emph{sma10}            & Simple $n$(10 \emph{here})-days Moving Average    & $C_t/\frac{C_t+C_{t-1}+...+C_{t-n}}{n}, n=10$\\
    \emph{slow\_k}    & Stochastic K\%      & $\frac{C_t-LL_{t-(n-1)}}{HH_{t-(n-1)}-LL_{t-(n-1)}}\times100$\\
    \emph{slow\_d}    & Stochastic D\%      &                           $\frac{\sum_{i=0}^{n-1}K_{t-i}}{10}\%$\\
    \emph{willr}   &   Larry William’s R\% &   $\frac{H_n-C_t}{H_n}-L_n \times100$\\
    \emph{macd\_diff}     & Moving Average Convergence Divergence (MACD) & $\frac{MACD(t)-MACD(t-1)}{C_t}$\\
    \emph{cci} &   Commodity Channel Index &   $\frac{M_t-SM_t}{0.015D_t}$\\
    \emph{rsi}     &   Relative Strength Index &   $100-\frac{100}{1}+(\sum_{{i=0}^{n-1}UP_{t-i}/n)/(\sum_{i=0}^{n-1}DW_{t-i}/n)}$\\
    \emph{adosc}   &   A/D (Accumulation/Distribution) Oscillator  & $AD(t)=AD(t-1)+CMFV(t)$\\
  \bottomrule
\end{tabular}
\end{center}
\begin{tablenotes}
\item$P_{long}$ and $P_{short}$ is the long position and short position currently holding respectively. $A$ is the total assets currently holding including stock and cash. $C_t$ is the closing price, $L_t$ is the low price and $H_t$ is the high price at time $t$. $LL_t$ and $HH_t$ implies lowest low and highest high in the last t days, respectively. $MACD(t)=DIFF(t)-EMA(t)_9$, where $EMA(t)_n=EMA(t-1)\times\frac{n-1}+C_\times\frac1n$ is the exponential moving average. $M_t=\frac{H_t+L_t+C_t}{3}, SM_t=\frac{\sum_{i=1}^{n}M_{t-i+1}}{n}, D_t=\frac{\sum_{i=1}^{n}|M_{t-i+1-SMt}|}{n}.$$UP_t$ is upward price change while $DW_t$ is downward price change at time $t$. $CMFV(t)=\frac{H_t-C_{t-1}}{H_t-Lt}*V_t$, where $V_t$ is the volume at time $t$.  
\end{tablenotes}
\end{table*}

\subsubsection{Output}
The model's output includes three actions: \emph{buy}, \emph{sell}, and volume. If the value of the \emph{buy} or \emph{sell} action is greater than the threshold of 0.5, the model will execute the corresponding action. If both the \emph{buy} and \emph{sell} actions have values greater than the threshold, the model will execute the action with the larger value. The \emph{volume} action indicates the volume of the order in relation to the total assets of the model. It is used to specify the size of the order being placed. This can be useful in managing risk, as it allows the model to adjust the size of its positions based on the level of risk it is willing to take on.

\subsection{Framework}

The project is implemented using Python and leverages several libraries and APIs. The NEAT-Python library\cite{neat} is used to build a NEAT  algorithm. The sqlite3 API\cite{sqlite} is used to manage the project's dataset, which is mentioned in section \textbf{Research data}.

During each generation of the NEAT algorithm, a portion of the train data is randomly extracted from the database and fed into the backtesting.py module. This module runs a backtest on each individual in the current generation, using the train data as input. The backtesting.py module provides a report of the results of the backtest, which is then used by a fitness function to evaluate the performance of each individual. The fitness function is used to determine which individuals should be selected to move on to the next generation and which should be discarded. 

\subsection{Fitness Function Design}
It's important to carefully design the fitness function in a NEAT algorithm, as it plays a crucial role in guiding the evolution of the population towards the desired objective.
\subsubsection{Option 1}

Our initial attempt is to use the System Quality Number (SQN) as the fitness function for our NEAT algorithm. 
\begin{equation}
    Fitness_1(R) = SQN = \#\ trades * Avg(PnL) /Std(PnL)
\end{equation}

Where:
\begin{itemize}
    \item $R$ is the backtest report produce by backtesting.py module.
    \item $\#\ trades$ is the total number of trades made by the trading system.
    \item $Avg(PnL)$ is the average profit and loss (PnL) of all transactions.
    \item $Std(PnL)$ is a measure of the dispersion of PnL values around the mean.
\end{itemize}

We found that the model developed a strategy of buying at the close price and holding until the end of the backtest. However, this strategy did not produce the desired results and did not achieve our project goals. As a result, we may need to revise the fitness function in order to more effectively guide the evolution of the population towards our desired objectives.

\subsubsection{Option 2}

In our second attempt to design the fitness function for the NEAT algorithm, we formalized our project goals into a multi-objective function. Our project goals include earning maximization, risk avoidance, and outperforming the Buy \& Hold strategy. To achieve these goals, we designed the fitness function to include three different metrics: PnL, PnL relative to the Buy \& Hold strategy, and the maximum drawdown of the capacity curve. The PnL metric represents the profit and loss of all transactions and aims to maximize the final earning. The PnL relative to the Buy \& Hold strategy metric represents the return relative to the Buy \& Hold strategy and aims to outperform this strategy. The maximum drawdown of the capacity curve metric represents the maximum drawdown of capacity during the backtest and aims to avoid risk. Our fitness function aims to optimize these multiple objectives simultaneously in order to achieve the desired outcomes of our project.
\begin{equation}
    Fitness_2(R) = PnL+1.5\times PnL_{relative}-0.5\times max(drawdown)
\end{equation}

Where:
\begin{itemize}
    \item $PnL$ is the profit and lose.
    \item $PnL_{relative}$ is the profit and lose relative to the Buy \& Hold strategy.
    \item $max(drawdown)$ is the maximum drawdown of the capacity curve.
\end{itemize}

In this case, the model developed a strategy similar to the Buy \& Hold strategy, which involves buying stocks and holding onto them until the backtest is over. While this strategy may have a high win rate due to the long-term uptrend of the stock market, it may not necessarily achieve the goals of the project. The Buy \& Hold strategy may result in a high capacity utilization rate, meaning that a large portion of the available capital is tied up in a small number of stocks, which can increase the risk exposure of the portfolio. Additionally, the Buy \& Hold strategy may not be the most effective way to achieve the project's goals, which may involve more active trading in order to maximize earnings and minimize risk. Therefore, it may be necessary to revise the fitness function in order to better align with the project's objectives.

\subsubsection{Option 3}
The final attempt is derived from the second one. To achieve the goals, we have designed a fitness function that rewards the agent for making trades and penalizes it for holding onto assets for extended periods of time. Specifically, we have added a reward for the number of trades made and a penalty for the average hold duration. By adjusting these metrics in the fitness function, we hope to encourage the agent to make profitable trades while minimizing the risk of long-term losses.
    \begin{equation}
\begin{aligned}
Fitness_3(R) = & PnL+1.5\times PnL_{relative}-0.5\times max(drawdown)+\\
    & 0.0005\times\#\ trades-avg(duration)
\end{aligned}
\end{equation}

Where:
\begin{itemize}
    \item $\#\ trades$ is the number of trades in the backtest.
    \item $avg(duration)$ is the average of holding duration of all trades.
\end{itemize}

This fitness function has been successful in training a model that is able to meet the project goals. Specifically, the model has learned to perform swing trades, which are trades that are held for a relatively short period of time and aim to profit from both uptrends and downtrends in the market.

\subsection{Efficiency}

 The NEAT algorithm involves the evaluation and evolution of many neural network models in a population, which can be time-consuming. To evaluate the performance of these models, a method called backtesting is used, which involves generating train data randomly from a database and evaluating the model's performance on this data. The longer the date range of each train data set and the more frequently the data is generated, the more time-consuming the training process becomes. 
 To improve the efficiency of the training process, we proposed a technique called progressive train data has been proposed. With this technique, shorter data sets are used for model evaluation in the earlier stages of training and the date range is progressively increased as training progresses. This helps to reduce the time complexity of the training process and allows the model to focus on different objectives at different stages of training. In our case, the model is trained on 90-days data in the first 1500 generations to focus on identifying reversals of trends and maximizing total return. In the next 400 generations, the model is trained on 150-days data, and in the final 100 generations, the model is trained on 1-year data to focus more on capacity and risk management to ensure long-term earning.

\section{Result and Evaluation}
After training the model using the NEAT algorithm and a multi-objective fitness function, we selected the best individual from the population for further evaluation. To do this, we performed backtests on each individual using random 365-day data for 20 times and compared their performance. When selecting the optimal individual, we considered several factors, including the average trade duration, the number of trades, and the System Quality Number (SQN). The ideal average trade duration is less than 90 days, and the number of trades should not be too high or too low. We also aimed to select the individual with the highest SQN, which is a measure of the quality of a trading system that takes into account the system's reliability, availability, maintainability, and performance. In this example, the individual with id 23554 was selected for further evaluation.

\begin{figure}
    \centering
    \includegraphics[height=!, width=\linewidth, keepaspectratio]{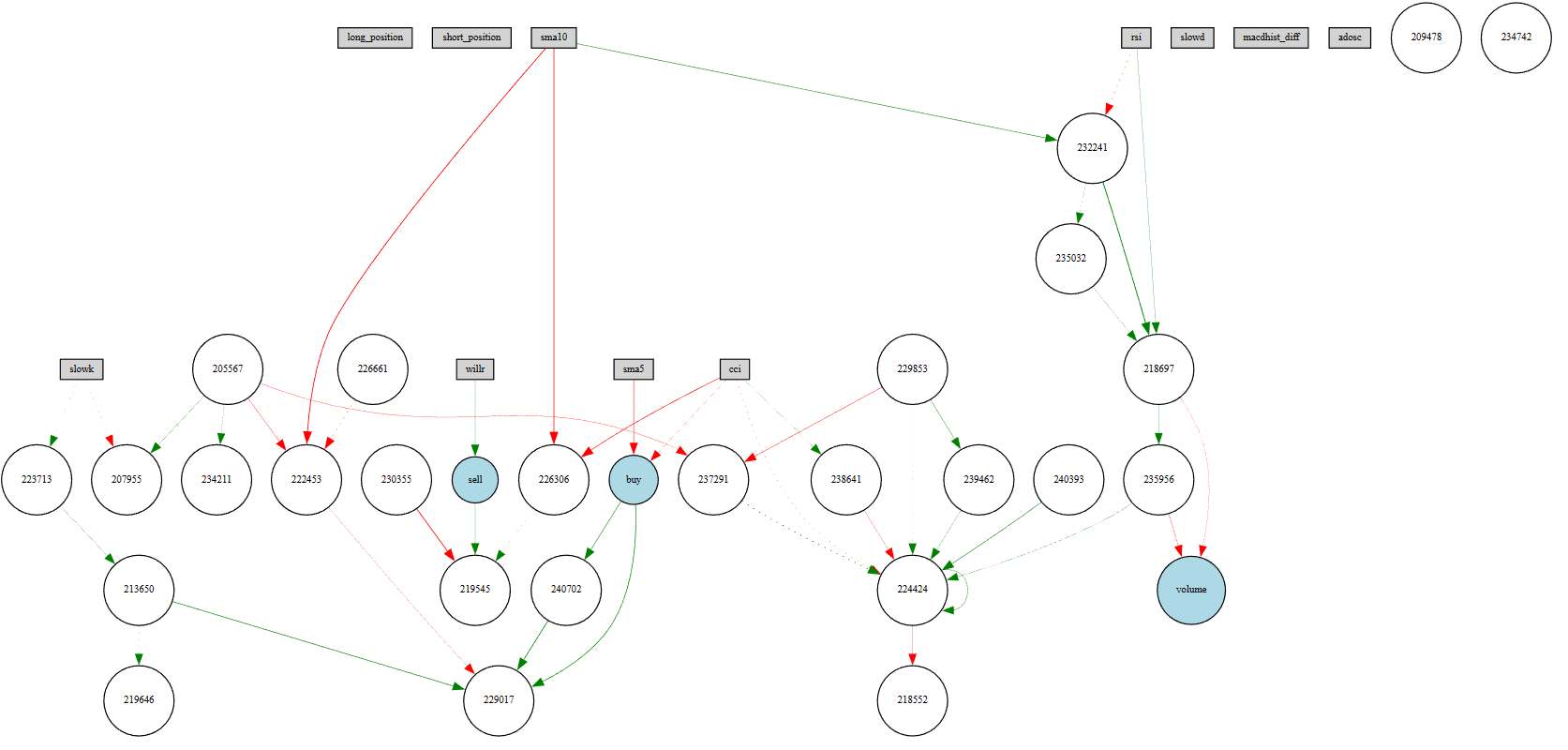}
    \caption{The example model architecture in the well-trained population.}
    \label{fig: model}
\end{figure}

The neural network architecture \ref{fig: model} of the model 23554 exhibits certain characteristics that influence its trading behavior. The model tends to buy when the price is high relative to the 5-day simple moving average and the commodity channel index (CCI) is high. On the other hand, the model tends to sell when the Larry Williams R\% is low. The volume of each action is controlled by the relative strength index (RSI) and the 10-day simple moving average. These characteristics may impact the model's performance, as they determine the timing and volume of its trades.

\begin{figure}
    \centering
    \includegraphics[height=!, width=\linewidth, keepaspectratio]{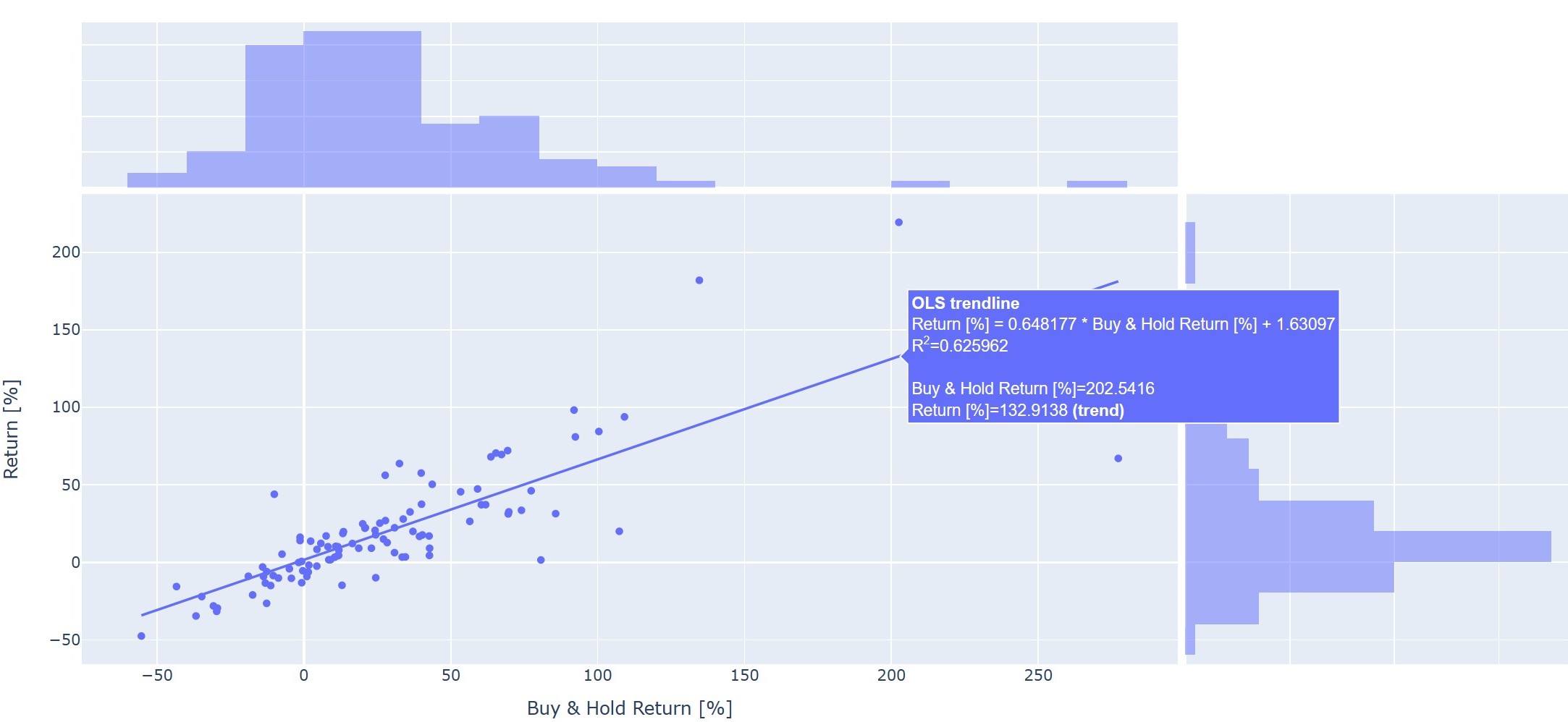}
    \caption{The comparison of model's strategy and Buy \& Hold strategy.}
    \label{fig: comparison}
\end{figure}

To more precisely evaluate the selected model, we performed additional backtests on it using random 1-year data for 100 times. We compared the performance \ref{fig: comparison} of the model's strategy to the Buy \& Hold strategy and observed that the model generally preferred to do long trades rather than short trades, as the return on the model's strategy was higher when the return on the Buy \& Hold strategy was higher. The overall trend line of the model's strategy had a positive correlation with the trend line of the Buy \& Hold strategy, but with an absent value lower than 1. This indicates that the model's strategy had a lower variance in average return and was more stable compared to the Buy \& Hold strategy. The table \ref{tab:performance} showed that the standard deviation of return for the model's strategy was 20\% lower than the value for the Buy \& Hold strategy. While the average return for the model's strategy was lower, it had approximately the same win rate and lower risk exposure due to the shorter exposure time.
\begin{table}[]
    \centering
    \begin{tabular}{ccl}
    \toprule
     Metrics    & Model & Buy \& Hold \\
    \midrule
         Average Return & $18.76\%$ & $27.97\%$\\
         Std of Return & $38.86\%$ & $47.73\%$\\
         Win Rate & $989/1402=71\%$ & $73/100=73\%$\\
         Relative Win Rate & $38$ & $62$\\
         Exposure Time & $86.98\%$ & $100\%$\\
    \bottomrule
    \end{tabular}
    \caption{The performance comparison of our model's strategy and Buy \& Hold strategy.}
    \label{tab:performance}
\end{table}

\section{Conclusion}
In this project, we developed a methodology to apply the NEAT algorithm to stock trading using multiple technical indicators. Through progressive training and a multi-objective fitness function, we were able to achieve similar returns to the Buy \& Hold strategy, while also achieving lower risk exposure and greater stability.

One of the key findings of this project was the importance of carefully designing the fitness function in order to guide the evolution of the population towards the desired objectives. By incorporating metrics that encouraged more active trading and avoided long-term holding, we were able to improve the performance and stability of the model.

However, we also encountered some challenges during the training process, including the presence of a large number of nodes, connections, and inputs that were not used in decision-making. This was a common issue in the NEAT algorithm and could potentially impact performance and efficiency. In the future, it may be worthwhile to explore ways to improve the algorithm in order to address this issue.

Overall, the results of this project have implications for practitioners in the field of stock trading, as they demonstrate the potential benefits of using the NEAT algorithm and technical indicators to develop trading strategies. However, there is still room for further exploration and improvement, particularly in terms of refining the model and the methodology used. In the future, it may be interesting to apply the same algorithm and framework to shorter interval data in order to assess the potential impact on performance.
\bibliographystyle{acm}
\bibliography{reference}
\end{document}